\pdfoutput=1

\documentclass[11pt]{article}

\usepackage[final]{acl}

\usepackage{times}
\usepackage{latexsym}

\usepackage[T1]{fontenc}

\usepackage[utf8]{inputenc}

\usepackage{microtype}

\usepackage{inconsolata}

\usepackage{graphicx}
\usepackage{enumitem}
\usepackage{kotex}
\usepackage{multirow}
\usepackage{booktabs}
\usepackage{placeins}
\usepackage{colortbl} 

\newcommand{\eg}{{\it e.g.}}
{}
\title{Transparent Reference-free Automated Evaluation of Open-Ended User Survey Responses}

\author{
    Subin An\textsuperscript{1}\thanks{Equal contribution. The authors are listed in alphabetical order.} \quad
    Yugyeong Ji\textsuperscript{2}\footnotemark[1] \quad
    Junyoung Kim\textsuperscript{2}\footnotemark[1] \quad
    Heejin Kook\textsuperscript{2}\footnotemark[1] \quad
    Yang Lu\textsuperscript{3}\thanks{Corresponding authors} \quad
    Josh Seltzer\textsuperscript{3}\footnotemark[2]\\
    \textsuperscript{1}Kookmin University \quad
    \textsuperscript{2}Sungkyunkwan University \quad
    \textsuperscript{3}Nexxt Intelligence \\
    \texttt{nesquiq@kookmin.ac.kr} \\
    \texttt{\{ygang29, junyoung44, hjkook\}@skku.edu} \\
    \texttt{\{yang, josh\}@nexxt.in}
}

\begin{document}
\maketitle

\begin{abstract}
Open-ended survey responses provide valuable insights in marketing research, but low-quality responses not only burden researchers with manual filtering but also risk leading to misleading conclusions, underscoring the need for effective evaluation. Existing automatic evaluation methods target LLM-generated text and inadequately assess human-written responses with their distinct characteristics. To address such characteristics, we propose a two-stage evaluation framework specifically designed for human survey responses. First, gibberish filtering removes nonsensical responses. Then, three dimensions—\textit{effort}, \textit{relevance}, and \textit{completeness}—are evaluated using LLM capabilities, grounded in empirical analysis of real-world survey data. Validation on English and Korean datasets shows that our framework not only outperforms existing metrics but also demonstrates high practical applicability for real-world applications such as response quality prediction and response rejection, showing strong correlations with expert assessment.

\end{abstract}
\section{Introduction}
\label{sec:introduction}

User surveys play a crucial role in marketing research, informing strategy and product development decisions~\cite{Crick2023, netzer2012mine}. Among these, open-ended responses are particularly valuable, as they offer rich insights through first-hand experiences and nuanced perspectives~\cite{Rouder2021What}. 
However, the quality of such responses can vary widely, and low-quality responses risk introducing noise or misleading interpretations. Moreover, identifying low-quality responses presents an opportunity to re-engage participants and elicit more meaningful feedback.

Evaluating the quality of open-ended responses is therefore essential, yet it presents two major challenges. First, manual evaluation of all responses by human annotators is prohibitively costly, necessitating automated evaluation methods. Second, the absence of ground-truth reference for comparison renders traditional reference-based metrics (\eg~BLEU~\cite{BLEU2002}, ROUGE~\cite{ROUGE2004}) inapplicable.

\begin{figure}[t]
\centering
\includegraphics[width=1.0\linewidth]{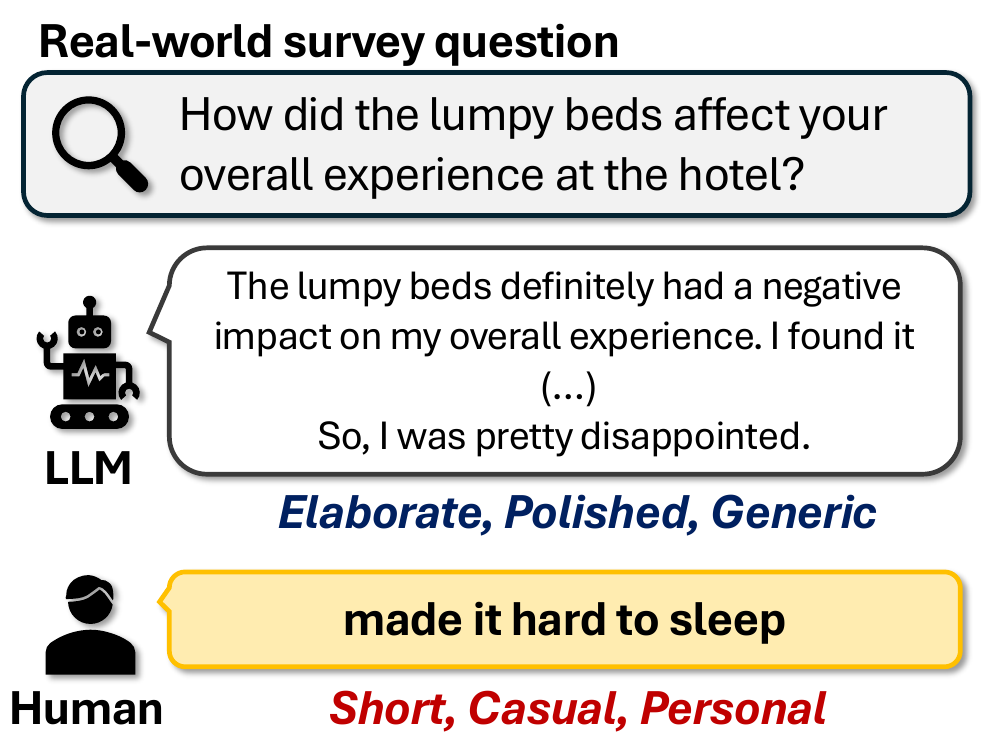} 
\caption{Example highlighting how human-written responses differ from LLM-generated ones.}
\label{fig:motivating}
\end{figure}

To address these issues, recent efforts have explored reference-free evaluation using large language models (LLMs) as judges~\cite{GEval2023, SeeEval2025, TALE2025, HD-Eval2024}.
These approaches leverage LLMs’ general language understanding to assess response quality without relying on predefined answers. Existing evaluation methods are primarily designed to assess responses generated by LLMs~\cite{LLMasJudge2023, AlternativeLLM2023}. However, as illustrated in Figure~\ref{fig:motivating}, human-written responses differ substantially in nature. They tend to be shorter, more careless, and often reflect personal or subjective perspectives. These differences highlight the need for evaluation criteria that account for the unique characteristics of human responses, rather than applying standards optimized for LLM-generated text.

In this paper, we propose a novel evaluation framework to address an industrial problem that has been overlooked by existing LLM evaluation studies: the evaluation of short, informal, and often noisy real human-written survey responses. 
Our framework is specifically designed for these unique characteristics, grounded in analysis of real-world user data. 
It incorporates a gibberish filtering module and introduces three novel evaluation dimensions—\textit{effort}, \textit{relevance}, and \textit{completeness}—to effectively assess the quality of human-written open-ended responses.
To demonstrate the practical utility and real-world applicability of our framework, we show how its scores can be leveraged for two key downstream applications: response quality prediction for analytical purposes and response rejection for operational quality control. These applications enable systems to enable researchers to gain more meaningful insights by assessing user responses, while also offering a mechanism to filter low-quality input and provide feedback to participants.
Furthermore, to evaluate its robustness across languages and cultural contexts, we validate the framework on both English and Korean datasets.

Our key contributions are as follows:
\begin{itemize}[leftmargin=5mm]
\vspace{-2mm}
\item We conduct a comprehensive analysis of real user responses, highlighting the need for evaluation criteria beyond conventional LLM-as-a-judge.
\vspace{-2mm}
\item We propose an interpretable, multi-dimensional evaluation framework tailored for open-ended responses, capturing the aspects of \textit{effort}, \textit{relevance}, and \textit{completeness}.
\vspace{-2mm}
\item We empirically validate our approach, showing strong performance in fine-grained dimension scoring and downstream applications like overall quality prediction and response rejection across English and Korean datasets.

\end{itemize}

\section{Related Work}
\label{sec:related_work}

\subsection{Reference-Free Evaluation Approaches}
Traditional evaluation methods have primarily relied on reference-based similarity measures such as BLEU~\cite{BLEU2002} and ROUGE~\cite{ROUGE2004}.
However, these approaches are unsuitable for open-ended, user-generated texts such as survey responses, where ground-truth reference often unavailable or are impractical to collect.
Moreover, they remain highly sensitive to the quality and completeness of reference data~\cite{BLEULimitation2020}.
This limitation has motivated reference-free evaluation methods that assess response quality without requiring references.

Among these, embedding-based approaches such as BERTScore~\cite{BERTScore2020} successfully evaluate semantic similarity but often fail to capture fine-grained aspects such as informativeness and logical coherence.
More recent frameworks introduce multi-dimensional evaluation by decomposing quality into distinct criteria, such as relevance and coherence~\cite{ConfuseLLM2024, HD-Eval2024}.
These approaches enable more interpretable and robust assessments, especially for open-ended tasks where constructing references is costly or infeasible~\cite{RefFreeSurvey2025}.

\subsection{LLM-Based Automatic Evaluation}

LLM-based evaluation methods have recently emerged as a compelling alternative for assessing natural language quality~\cite{LLMasJudgeSurvey2024}.
This shift is driven by the cost and limited scalability of human evaluation, as well as the inability of metrics such as BLEU~\cite{BLEU2002}, ROUGE~\cite{ROUGE2004}, and BERTScore~\cite{BERTScore2020} to capture subtle nuance.
In contrast, LLMs provide rich language understanding and reasoning, enabling multi-dimensional assessment even without references.
For example, G-Eval~\cite{GEval2023} uses LLMs like GPT-4 to produce quantitative, interpretable scores from natural language prompts under predefined criteria.
This approach overcomes the limitations of similarity-based metrics that often fail to reflect semantic coherence and logical consistency.




\begin{figure}[t]
\centering
\includegraphics[width=1.0\linewidth]{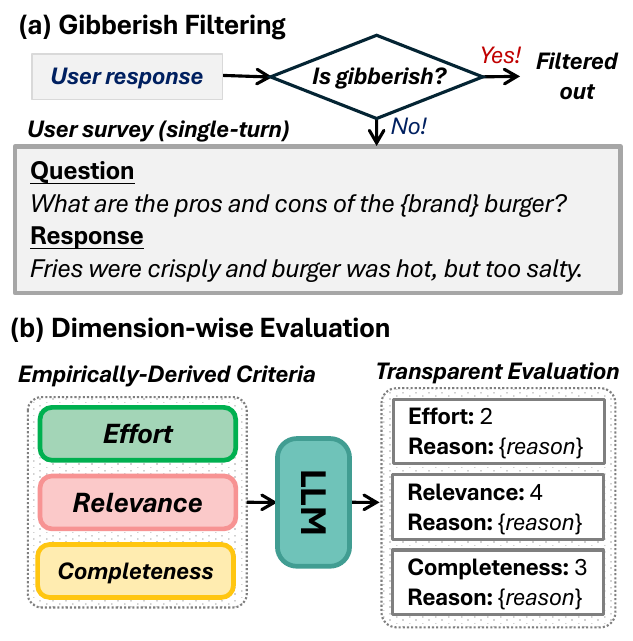} 
\caption{Overall pipeline of the user response evaluation, showing gibberish filtering and dimension-wise evaluation steps.}
\label{fig:main}
\end{figure}

\section{User Response Evaluation}

We design a structured framework to evaluate the quality of user-generated open-ended responses.
Figure~\ref{fig:main} illustrates the overview of the proposed framework.
The evaluation consists of two main stages: (1) gibberish filtering, which removes nonsensical or malformed inputs, and (2) dimension-wise scoring based on three core criteria—\textit{effort}, \textit{relevance}, and \textit{completeness}. 
Each dimension is evaluated independently using a prompt-based inference with LLMs, with scoring rubrics and qualitative reasoning to ensure interpretability and consistency. 
In the following, we detail the rationale, presented in the Preliminary Study, and the scoring method, detailed in the Method section, for each dimension.

\subsection{Gibberish Filtering}
\label{gibberish}
Gibberish, which refers to meaningless sequences or nonsensical language (\eg, asdf, ddddd), is a phenomenon that rarely appears in LLM-generated responses but frequently occurs in real-world human inputs. 
Detecting such gibberish in user responses is crucial for reducing the computational cost of LLM-based evaluation by preemptively filtering out samples that lack informative content. 
By explicitly addressing noisy, real-world inputs which are underrepresented in curated LLM outputs, our framework not only complements but, in key settings, also surpasses language-agnostic evaluators in applicability.
Unlike generic filtering methods, our approach is tailored to the structural and statistical properties of individual languages. 
We propose language-specific detection pipelines for English and Korean that incorporate both statistical modeling and linguistic heuristics,thereby capturing the diverse manifestations of gibberish across languages. 
For details on the detection architecture, refer to Appendix~\ref{appendix:gibberish}.

\subsection{Dimension-wise Evaluation}
\label{dimension_eval_method}
Existing LLM-based evaluators, designed for model outputs, often fail to capture the unique characteristics of human responses. To address this challenge, we conducted a data-driven criteria establishment process. By systematically analyzing large-scale real-world English and Korean survey data, we empirically identified \textit{effort}, \textit{relevance}, and \textit{completeness} as the core dimensions. 

Based on these findings, we develop new evaluation criteria tailored to human-generated responses. Our method uses a simple prompting approach grounded in these empirical insights, enabling effective evaluation of real-world survey responses in a manner aligned with how marketing researchers assess quality. The LLM assigns a score and provides a brief justification based on these criteria. Full prompts are provided in Appendix~\ref{appendix:prompts}.

\vspace{-2mm}

\subsubsection{Effort}
\paragraph{Preliminary Study}
In previous LLM-based response evaluations, the \textit{effort} dimension was often considered unnecessary, as LLMs do not exhibit human-like “effort.” However, since real survey respondents are human, assessing how sincerely they answer questions is crucial~\cite{krosnick1991response}.

One challenge is that human responses are often brief, which can easily be mistaken for a lack of effort~\cite{santilli2025revisitinguncertaintyquantificationevaluation}. Yet short responses can still show high effort if they provide enough information and specificity. Conversely, length alone does not guarantee thoughtfulness.

Evaluating effort in human responses therefore requires criteria that go beyond length or wording, taking into account informational density and specificity. This study presents an evaluation approach designed with these factors in mind, making it more suitable for real user responses and aligned with how marketing researchers assess response quality.

\paragraph{Method}
The \textit{effort} dimension evaluates the level of cognitive effort demonstrated in the user response, with a focus on two aspects: \textit{informativeness} and \textit{specificity}. The LLM assigns a discrete score on a 0–7 scale, where higher scores reflect more detailed, thoughtful, and specific responses. A rubric was designed to define clear criteria for each score range. The model is prompted with this rubric and returns both the score and a natural language justification.

\begin{figure}[t]
\centering
\includegraphics[width=1.0\linewidth]{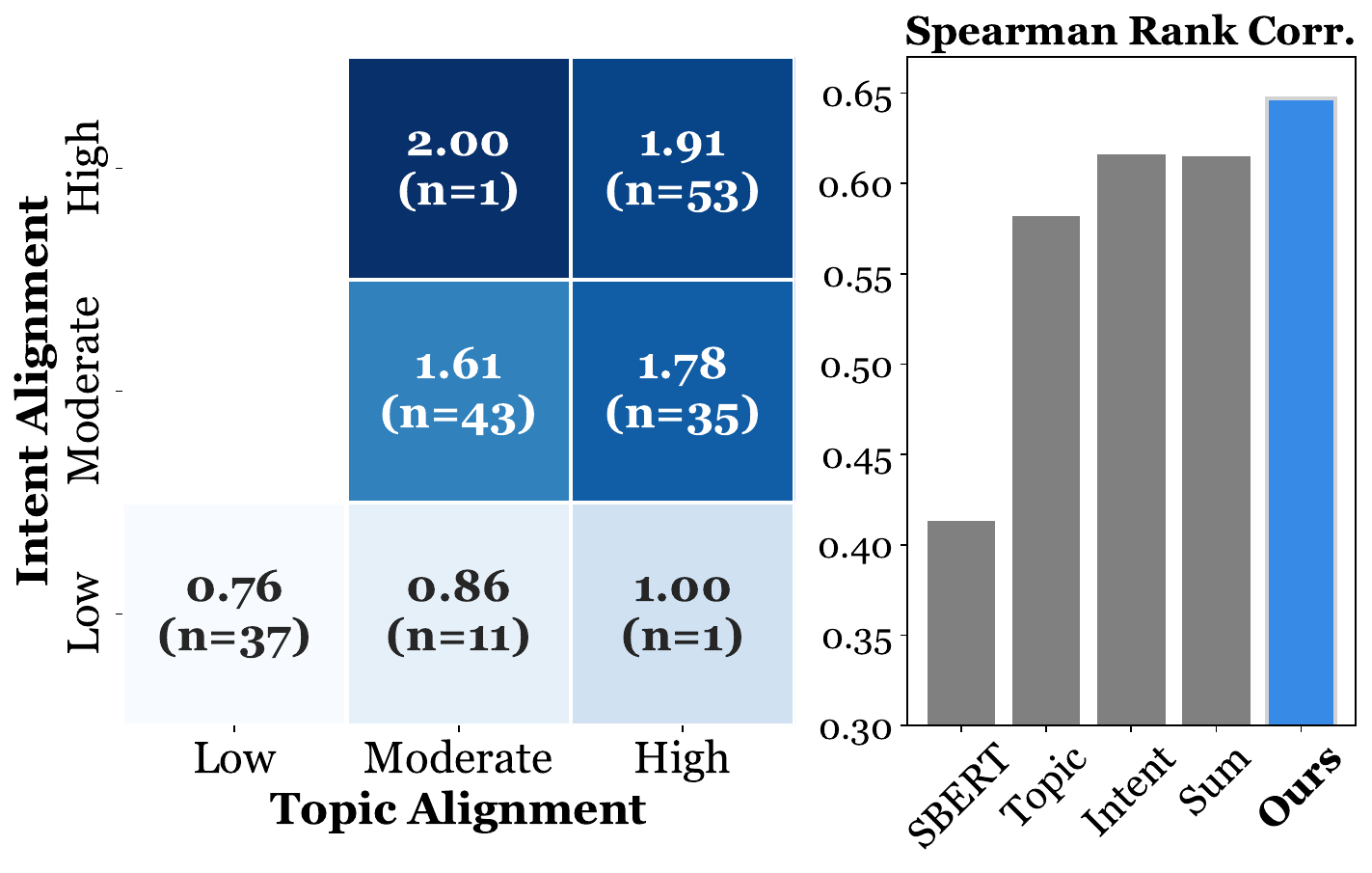} 
\caption{Heatmap (left) showing human expert-rated relevance scores by LLM-evaluated topic alignment (x-axis) and intent alignment (y-axis). Bar chart (right) shows the Spearman rank correlation of our proposed method (“Ours”) and baseline methods.}
\label{fig:preliminary_relevance}
\end{figure}

\subsubsection{Relevance}
\paragraph{Preliminary Study}

In survey response evaluation, relevance is generally assessed based on both topic alignment and intent alignment. In other words, a response must not only address the topic of the question but also faithfully reflect its intent.

For open-ended survey responses, which often incorporate the respondent’s unique experiences and context, simple embedding-based similarity measures are insufficient to fully capture this alignment~\cite{FAN2023}. Addressing this limitation requires the high-level language understanding and inference capabilities of LLMs to grasp the implicit meaning behind questions.

Figure~\ref{fig:preliminary_relevance} shows how we used an LLM to evaluate both the topic and intent alignment of responses and compares these evaluations to human judgments. The results indicate that responses reflecting the question’s intent, in addition to being topically relevant, were rated as having higher relevance overall. The approach that jointly considered both topic and intent achieved the highest agreement with human judgments. This suggests that inference-based evaluation using LLMs provides a more nuanced and human-aligned assessment compared to simple similarity-based methods.

\paragraph{Method}
The \textit{relevance} dimension measures how well a response aligns with both the topic and the intent of the given question. It is assessed on a discrete scale from 0 to 4, ranging from completely irrelevant to fully relevant. Each level is defined based on two criteria: topic alignment and intent alignment.

\begin{figure}[t]
\centering
\includegraphics[width=1.0\linewidth]{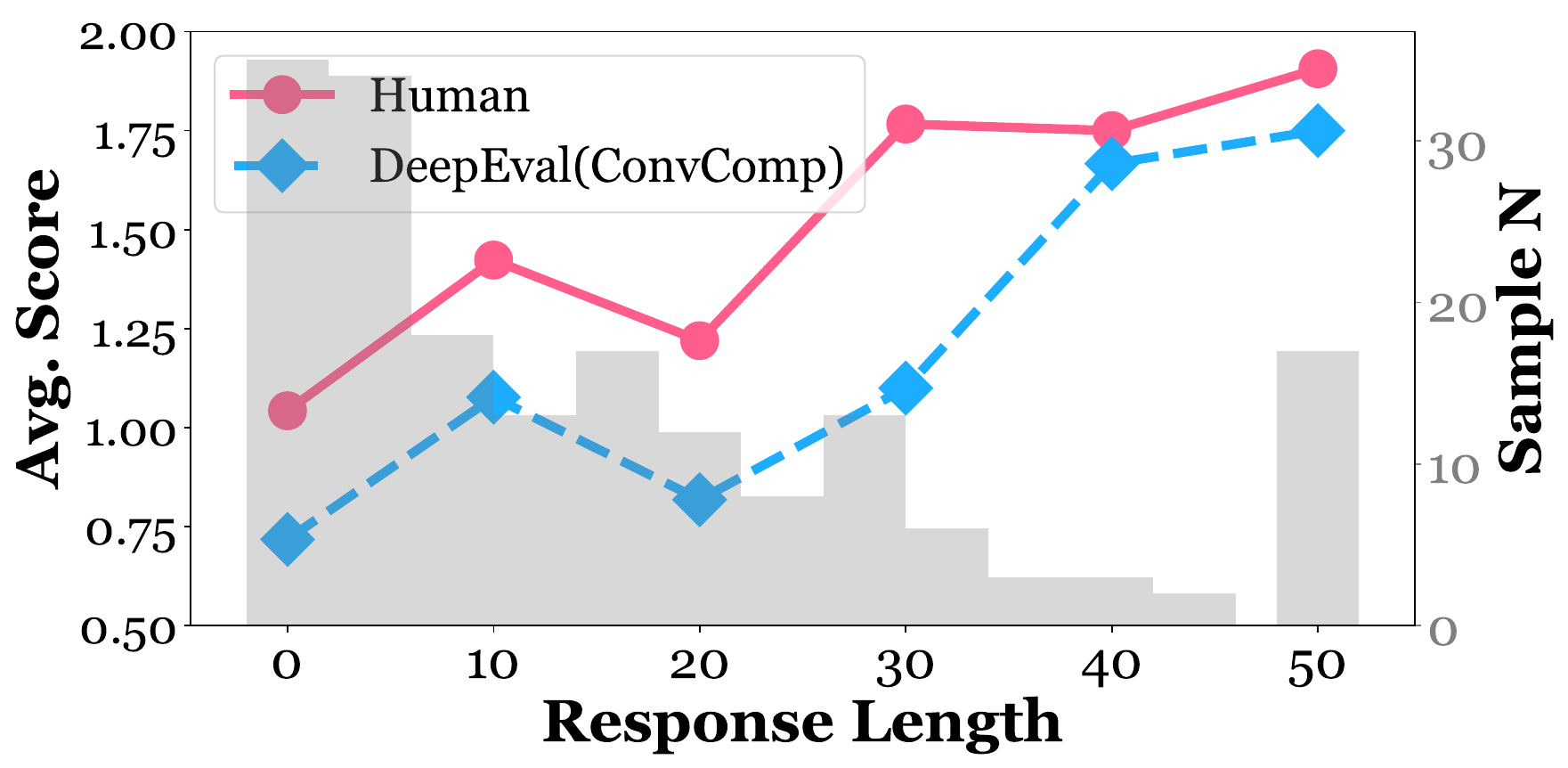} 
\caption{Average human expert-rated completeness scores (pink) and DeepEval Conversational Completeness scores (blue) across different response length bins. Grey bars indicate sample counts per bin. }
\label{fig:preliminary_completeness}
\end{figure}

\subsubsection{Completeness}
\paragraph{Preliminary Study} 
User-written open-ended responses are often brief yet sufficient, conveying key information without explanation. 
This concise style is natural in human communication but often undervalued by automatic evaluation methods that rely on surface-level features such as length or keyword count.
As a result, verbose but uninformative responses are often overrated, while concise yet complete reponses are penalized.
Figure~\ref{fig:preliminary_completeness} shows the distribution of human-written response lengths and how human and LLM evaluations differ in treating length.

We redefine \textit{completeness} as the extent to which a response fulfills the core informational intent of the question. 
A complete response addresses all essential elements—explicit or implied—needed to satisfy what the question fundamentally asks. 
Partial responses missing key components or failing to engage with the question's intent are rated lower, regardless of length.
To avoid redundancy, elaboration and contextual fit are excluded from this dimension and evaluated under \textit{effort} and \textit{relevance}, as suggested in prior work~\cite{ConfuseLLM2024}.

\paragraph{Method}
The \textit{completeness} dimension is evaluated on a discrete 0–4 scale, mapped onto three qualitative categories: \textit{Not Fulfilled}, \textit{Partially Fulfilled}, and \textit{Fully Fulfilled}. 
The LLM receives a tailored prompt containing the question, user response, and scoring rubric. 
The rubric guides the model to assess whether the response adequately covers the informational requirements of the question.

\section{Experimental Settings}
\label{sec:settings}

\begin{table}[]
\small
\centering
\begin{tabular}{l|cc|cc}
\toprule
\textbf{Language}     & \multicolumn{2}{c|}{\textbf{English}}                                 & \multicolumn{2}{c}{\textbf{Korean}}                                  \\ \midrule
\textbf{Dataset}      & \multicolumn{1}{r}{\textbf{Dev}} & \multicolumn{1}{r|}{\textbf{Test}} & \multicolumn{1}{r}{\textbf{Dev}} & \multicolumn{1}{r}{\textbf{Test}} \\ \midrule
\textbf{\# Response}  & \multicolumn{1}{r}{181}          & \multicolumn{1}{r|}{94}            & \multicolumn{1}{r}{122}          & \multicolumn{1}{r}{100}           \\
\textbf{Dim. Annot.}  & Crowd                         & Expert                             & Crowd                         & Crowd                          \\
\textbf{Prac. Annot.} & Crowd                         & Expert                             & Crowd                         & Expert                            \\ \bottomrule
\end{tabular}
\caption{Summary comparing English and Korean datasets across annotation aspects. "\# Response" indicates the number of dataset samples, while "Dim. Annot." and "Prac. Annot." refer to annotators for per-dimension annotations and practical usage (overall quality and response acceptance) annotations, respectively.}
\label{tab:dataset_summary_basic}
\end{table}

\begin{table*}[t]
\centering
\begin{tabular}{c c | c c | c c}
\toprule
\multicolumn{2}{c|}{\textbf{Language}} & \multicolumn{2}{c|}{\textbf{English}} & \multicolumn{2}{c}{\textbf{Korean}} \\
\cmidrule(r){3-4} \cmidrule(l){5-6}
\textbf{Dimension} & \textbf{Method} & Spearman $\rho$ & Kendall $\tau$ & Spearman $\rho$ & Kendall $\tau$ \\
\midrule

\multirow{2}{*}{Effort}
 & Length (token)   & 0.7889 & 0.6613 & 0.7621 & 0.6212 \\
 & \textbf{Ours}    & \textbf{0.8198} & \textbf{0.7223} & \textbf{0.7950} & \textbf{0.6797} \\
\midrule

\multirow{5}{*}{Relevance}
 & SBERT(all-MiniLM)      & 0.5839 & 0.4644 & 0.3429 & 0.2540 \\
 & SBERT(multi-lingual)   & 0.5223 & 0.4139 & 0.5538 & 0.4276 \\
 & DeepEval(AnswerRel)    & 0.5540 & 0.4759 & 0.4707 & 0.4191 \\
 & DeepEval(ConvRel)      & 0.4946 & 0.4597 & 0.2497 & 0.2307 \\
 & \textbf{Ours}          & \textbf{0.8614} & \textbf{0.7954} & \textbf{0.6021} & \textbf{0.5269} \\
\midrule

\multirow{3}{*}{Completeness}
 & Length (token)         & 0.6722 & 0.5374 & 0.6571 & 0.5171 \\
 & DeepEval(ConvComp)     & 0.4292 & 0.3859 & 0.3929 & 0.3478 \\
 & \textbf{Ours}          & \textbf{0.8245} & \textbf{0.7438} & \textbf{0.7457} & \textbf{0.6372} \\
\bottomrule
\end{tabular}

\caption{Spearman’s $\rho$ and Kendall’s $\tau$ correlations with human ratings for \textit{effort}, \textit{relevance}, and \textit{completeness} dimensions across English and Korean datasets. Further results on expert annotations, crowd annotations, and inter-annotator agreement can be found in Appendix~\ref{appendix:correlation_analysis}.}
\label{tab:main}
\end{table*}

\subsection{Datasets}
\label{sec:datasets}
We conducted our study using human-written open-ended survey responses in English and Korean, collected from real-world datasets provided by \textit{Nexxt Intelligence}. All responses were anonymized, and sensitive or personally identifiable information was thoroughly masked to protect participant confidentiality.
To ensure diversity, responses were sampled across question types, lengths, and quality levels. 

All responses were annotated for four core dimensions: \textit{effort}, \textit{relevance}, \textit{completeness} and \textit{gibberish}, as well as two dimensions of practical use: overall quality and response acceptance, by one to three crowd annotators (with averaged scores reported) or by a single domain expert, to assess how well our proposed method aligns with human judgments.
For expert annotation of Korean data, question–response pairs were carefully translated into English to preserve their original nuance as much as possible.
A summary of the dataset is shown in Table~\ref{tab:dataset_summary_basic}; details on annotation procedures and guidelines can be found in Appendix~\ref{appendix:dataset} and Appendix~\ref{appendix:human_annotation}.

\subsection{Evalaution}
\paragraph{Metrics}
To evaluate how well LLM scores align with human judgments, we compute Spearman (rank-based correlation)~\cite{spearman1987proof} and Kendall tau (ordinal association)~\cite{kendall1938new} correlations between the raw LLM scores and the average human annotations across the three evaluation dimensions: \textit{effort}, \textit{relevance}, and \textit{completeness}.

\paragraph{Baselines and Implementation Details}
We select baselines for each evaluation dimension. For Effort, since there is no existing LLM-based model for evaluating \textit{effort}, we use a heuristic method based on token length (\textbf{Length (token)}). For Relevance, we employ Sentence-BERT, specifically \textbf{SBERT(all-MiniLM)}\footnote{\url{https://huggingface.co/sentence-transformers/all-MiniLM-L6-v2}} and its multilingual variant \textbf{SBERT(multi-lingual)}\footnote{\url{https://huggingface.co/microsoft/Multilingual-MiniLM-L12-H384}}. For \textit{relevance} and \textit{completeness}, we use the default LLM-based automatic evaluation metrics provided by \textbf{DeepEval}\footnote{\url{https://deepeval.com/}}: AnswerRelevancy (AnswerRel), ConversationRelevancy (ConvRel), and ConversationalCompleteness (ConvComp). Both our proposed method and the LLM-based baselines were implemented using OpenAI's \texttt{gpt-4o-mini-2024-07-18} for inference, with the temperature set to 0 and the maximum tokens limited to 300.

\section{Results and Analysis}
\label{sec:results}

\subsection{Per-Dimension Performance}
We analyze how well the proposed method aligns with human annotations across the three evaluation dimensions—\textit{effort}, \textit{relevance}, and \textit{completeness}—in comparison to existing baselines.
We use Spearman’s rank correlation coefficient ($\rho$) and Kendall’s rank concordance coefficient ($\tau$) as evaluation metrics, computed separately for English (En) and Korean (Ko) responses.
Table~\ref{tab:main} presents the correlation results.

\paragraph{Effort}
For the \textit{effort} dimension, we evaluate the performance of a length-based baseline by measuring the correlation between the word count of each response and the corresponding annotated effort score.
The results show that response length shows moderate alignment with perceived \textit{effort} but fails to capture qualitative aspects like informational density.
In contrast, our prompt-based approach achieves consistently higher correlations in both English and Korean. These results suggest that content-aware evaluation is more precise and reliable than simple length-based measures for assessing \textit{effort}.

\paragraph{Relevance}
For the \textit{relevance} dimension, we compare our method with sentence embedding models (SBERT~\cite{SBERT2019}) and the LLM evaluation toolkit DeepEval~\cite{Ip_deepeval_2025}.
SBERT shows weak to moderate alignment with human annotations, with particularly low agreement in Korean. DeepEval Relevance metric performs comparably or slightly worse than SBERT, indicating limited effectiveness in capturing human notions of relevance.
In contrast, our prompt-based method shows the highest alignment with human ratings in both English and Korean, consistently outperforming all baselines. These results suggest that incorporating question topic and intent, rather than relying solely on surface-level similarity, yields more accurate and human-aligned \textit{relevance} assessments.

\paragraph{Completeness}
For the \textit{completeness} dimension, we compare our method against length-based and LLM-based evaluation baselines.
Both the length-based metric, computed as in the \textit{effort} dimension, and DeepEval’s conversation completeness metric show only moderate or poor alignment with human annotations.
In contrast, our prompt-based method consistently outperforms all baselines in both English and Korean.
These results suggest that our method, by accounting for not only response length and explicit requirements but also implicit cues and contextual completeness, provides a more comprehensive and reliable evaluation of \textit{completeness}.

\subsection{Practical Usage}
In this section, we evaluate how our framework's outputs can be applied to practical use cases. We aggregate the dimension-wise scores to demonstrate their utility in two key downstream applications: (i) \textbf{Response Quality Prediction}, where the aggregated score provides an interpretable, quantitative basis for researchers to analyze response quality, and (ii) \textbf{Response Rejection}, where the same score can be operationalized with a threshold to automate quality control. This evaluation is based on annotations from marketing researchers for both Korean and English responses.

\begin{table}[t]
\small
\centering
\renewcommand{\arraystretch}{1.2}
\setlength{\tabcolsep}{8pt}
\begin{tabular}{lcccc}
\toprule
\multirow{2}{*}{\textbf{Aggregation}} & \multicolumn{2}{c}{\textbf{English}} & \multicolumn{2}{c}{\textbf{Korean}} \\
\cmidrule(l{3pt}r{3pt}){2-3} \cmidrule(l{3pt}r{3pt}){4-5}
 & $\rho$ & $\tau$ & $\rho$ & $\tau$ \\
\midrule
Sum         & 0.90 & 0.78 & 0.60 & 0.51 \\
Regression  & 0.90 & 0.79 & 0.61 & 0.51 \\
LLM         & 0.87 & 0.79 & 0.54 & 0.50 \\
\bottomrule
\end{tabular}
\caption{Spearman’s $\rho$ and Kendall’s $\tau$ correlations with human ratings for overall quality prediction using different aggregation methods.}
\label{tab:overall_quality}
\end{table}

\vspace{-2mm}

\paragraph{Overall Quality}
\label{overall_quality_method}
Table~\ref{tab:overall_quality} shows the performance of three aggregation methods for combining dimension scores in both English and Korean datasets. After normalizing scores between 0 and 1, we aggregated them using Sum (simple addition), Regression (ridge regression-based weighted sum), and an LLM-based approach (prompt-based synthesis of scores and reasons). Correlations with expert ratings were evaluated for both languages.

All methods show consistent performance, with the LLM-based approach additionally providing reasoning and explanations to enhance transparency for marketing researchers. Cost-efficient methods such as Sum and Regression also performed strongly, suggesting practical alternatives depending on resource constraints. The LLM prompt is detailed in Table~\ref{tab:prompt_overall_quality}.

\begin{figure}[t]
\centering
\includegraphics[width=1.0\linewidth]{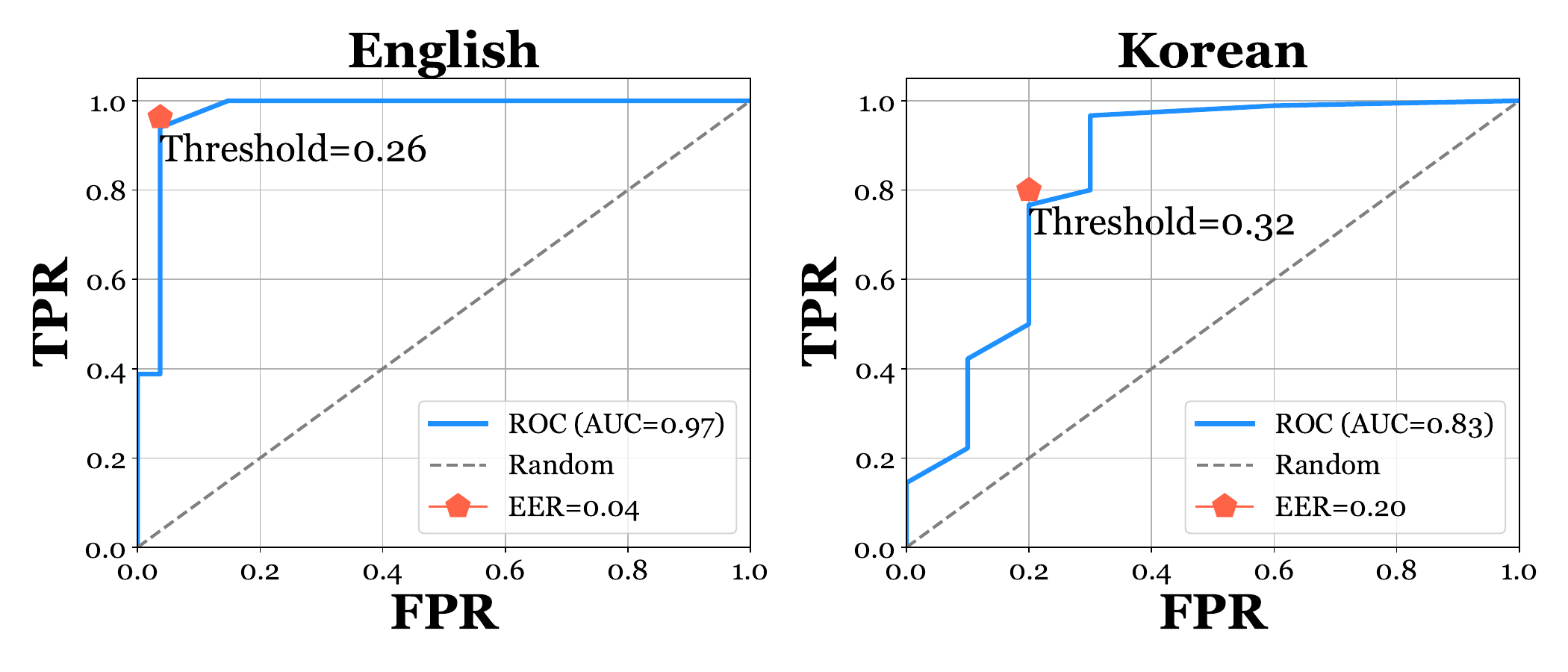} 
\caption{ROC curves for the English (left) and Korean (right) datasets.}
\label{fig:acceptance}
\end{figure}

\paragraph{Response Acceptance}

Figure~\ref{fig:acceptance} shows the ROC curves for acceptance classification for both English and Korean datasets. We computed overall quality using sum aggregation, normalized the scores between 0 and 1. Expert annotations were used as ground truth, where each response was labeled as `accept', `hold', or `reject'; `hold' and `reject' were treated as rejection in this evaluation.

The proposed method achieved an AUC of 0.97 for English and 0.83 for Korean, indicating strong alignment with expert judgments in both languages. However, the performance for Korean was lower, possibly due to meaning distortion or loss of nuance during translation into English for annotation.


\section{Future Works}
Our work serves as a foundation for building smarter, more trustworthy, and more versatile survey systems. By providing a reliable and interpretable framework for evaluating open-ended responses across \textit{effort}, \textit{relevance}, and \textit{completeness}, we set the stage for a series of impactful extensions:
\paragraph{Agentic and Real-Time Integration}
The framework can power multi-turn, agentic survey platforms that evaluate responses as they are given, flag shortcomings in specific dimensions, and generate targeted follow-up questions. This transforms surveys from static questionnaires into adaptive systems that improve data quality on the spot.
\paragraph{Safeguarding Data Quality}
Dimension-wise scores naturally lend themselves to detecting low-effort, insincere, or automated responses. Integrating such detection directly into survey workflows helps preserve validity, reduce downstream noise, and cut the costs of cleaning unusable data.
\paragraph{Cross-Lingual and Cross-Cultural Robustness}
Extending the framework beyond English and Korean ensures it generalizes across diverse populations. This not only supports broader applicability but also provides a basis for identifying and mitigating biases in multilingual and multicultural contexts.
\paragraph{Pathway to Quantitative Analysis}
The interpretability of dimension-wise scores offers a bridge to quantitative utilization. By aggregating or modeling these scores, researchers can derive higher-order insights and embed them into broader analytic pipelines.
In short, the immediate contribution of this work is methodological clarity and reliability. The long-term value lies in its adaptability: a foundation from which future research can build survey systems that are dynamic, bias-aware, and analytically powerful.

\section{Conclusion}
\label{sec:conclusion}
In this paper, we presented a two-stage evaluation framework tailored to human-written open-ended survey responses, combining gibberish detection with dimension-wise evaluation of \textit{effort}, \textit{relevance}, and \textit{completeness}. Our experiments demonstrated strong alignment with expert assessments in both English and Korean datasets, underscoring the framework’s cross-lingual applicability. These results highlight the potential of our method as an effective and scalable solution for supporting decision-making in industrial contexts. In future work, its interpretable outputs could be leveraged to construct targeted follow-up questions that supplement missing information and encourage user engagement, further enhancing its practical utility in real-world marketing research.


\section*{Limitations}

Although the proposed framework shows high concordance with expert judgments and consistently outperforms prior automated evaluation methods, it still presents several notable limitations.

First, the scoring process still depends on the subjective inference of large language models.
Despite using rubric-based prompts and examples to clarify evaluation criteria, final scores reflect the data distributions and stylistic preferences learned during pretraining.
While mitigation strategies were applied, such biases are difficult to fully eliminate due to the inherent characteristics of LLMs.

Second, the framework provides natural language explanations alongside each dimension score to enhance interpretability.
However, these explanations are ultimately generated outputs from the LLM and do not explicitly reveal the underlying rationale or token-level evidence used in scoring.
As a result, there are structural limitations in rigorously verifying the reliability or auditing the consistency of the evaluation outcomes.

Third, the three evaluation dimensions—\textit{effort}, \textit{relevance}, and \textit{completeness}—were introduced based on practical considerations, analyses of real user responses, and established evaluation practices in marketing research.
Experiments confirmed that these dimensions are effective in distinguishing key qualitative aspects of responses and may provide helpful guidance for subsequent applications, including identifying or supplementing incomplete information in user response.
Nonetheless, to establish these dimensions as general and theoretically grounded criteria for evaluating response quality, further discussion and systematic investigation are required.
Fourth, while data release is crucial for transparency and reproducibility, the dataset used in this study cannot be made publicly available. The data is proprietary to our company, contains confidential client information, and was collected without securing consent for public release. To provide transparency within these constraints, we describe the dataset details and labeling process in Appendix~\ref{appendix:dataset}.

\section*{Acknowledgments}

This work was supported by Institute of Information \& communications Technology Planning \&
Evaluation (IITP) grant funded by the Korea government(MSIT) (RS-2022-00143911, AI
Excellence Global Innovative Leader Education Program). The authors gratefully acknowledge Ellie Ma for her assistance with data annotation.

\bibliography{anthology,custom}

\begin{thebibliography}{26}
\providecommand{\natexlab}[1]{#1}

\bibitem[{Badshah et~al.(2025)Badshah, Emami, and Sajjad}]{TALE2025}
Sher Badshah, Ali Emami, and Hassan Sajjad. 2025.
\newblock \href {https://arxiv.org/abs/2504.07385} {Tale: A tool-augmented framework for reference-free evaluation of large language models}.
\newblock \emph{CoRR}.

\bibitem[{Bird et~al.(2009)Bird, Klein, and Loper}]{nltk@2009}
Steven Bird, Ewan Klein, and Edward Loper. 2009.
\newblock \href {http://www.oreilly.de/catalog/9780596516499/index.html} {\emph{Natural Language Processing with Python}}.
\newblock O'Reilly.

\bibitem[{Chiang and Lee(2023)}]{AlternativeLLM2023}
David~Cheng{-}Han Chiang and Hung{-}yi Lee. 2023.
\newblock \href {https://doi.org/10.18653/v1/2023.acl-long.870} {Can large language models be an alternative to human evaluations?}
\newblock In \emph{ACL}, pages 15607--15631.

\bibitem[{Crick(2024)}]{Crick2023}
James~M. Crick. 2024.
\newblock \href {https://doi.org/10.1080/0965254X.2023.2176533} {Analyzing survey data in marketing research: A guide for academics and postgraduate students}.
\newblock \emph{Journal of Strategic Marketing}, 32(2):203--215.

\bibitem[{Fan et~al.(2023)Fan, Aumiller, and Gertz}]{FAN2023}
Jing Fan, Dennis Aumiller, and Michael Gertz. 2023.
\newblock Evaluating factual consistency of texts with semantic role labeling.
\newblock In \emph{SEM}, pages 89--100.

\bibitem[{Freitag et~al.(2020)Freitag, Grangier, and Caswell}]{BLEULimitation2020}
Markus Freitag, David Grangier, and Isaac Caswell. 2020.
\newblock \href {https://doi.org/10.18653/v1/2020.emnlp-main.5} {{BLEU} might be guilty but references are not innocent}.
\newblock In \emph{EMNLP}, pages 61--71.

\bibitem[{Gu et~al.(2024)Gu, Jiang, Shi, Tan, Zhai, Xu, Li, Shen, Ma, Liu, Wang, and Guo}]{LLMasJudgeSurvey2024}
Jiawei Gu, Xuhui Jiang, Zhichao Shi, Hexiang Tan, Xuehao Zhai, Chengjin Xu, Wei Li, Yinghan Shen, Shengjie Ma, Honghao Liu, Yuanzhuo Wang, and Jian Guo. 2024.
\newblock \href {https://doi.org/10.48550/arXiv.2411.15594} {A survey on llm-as-a-judge}.
\newblock \emph{CoRR}.

\bibitem[{Hu et~al.(2024)Hu, Gao, Hu, Zhang, Chen, Xu, and Wan}]{ConfuseLLM2024}
Xinyu Hu, Mingqi Gao, Sen Hu, Yang Zhang, Yicheng Chen, Teng Xu, and Xiaojun Wan. 2024.
\newblock \href {https://arxiv.org/abs/2402.12055} {Are llm-based evaluators confusing nlg quality criteria?}
\newblock \emph{CoRR}.

\bibitem[{Ip and Vongthongsri(2025)}]{Ip_deepeval_2025}
Jeffrey Ip and Kritin Vongthongsri. 2025.
\newblock {deepeval}.
\newblock \url{https://github.com/confident-ai/deepeval}.
\newblock Apache-2.0 License. Accessed: 2025-06-26.

\bibitem[{Ito et~al.(2025)Ito, van Deemter, and Suzuki}]{RefFreeSurvey2025}
Takumi Ito, Kees van Deemter, and Jun Suzuki. 2025.
\newblock \href {https://arxiv.org/abs/2501.12011} {Reference-free evaluation metrics for text generation: A survey}.
\newblock \emph{CoRR}.

\bibitem[{Jindal(2021)}]{gibberishdetector@2021}
Madhur Jindal. 2021.
\newblock \href {https://huggingface.co/madhurjindal/autonlp-Gibberish-Detector-492513457} {Gibberish detector: High-accuracy text classification model}.

\bibitem[{Kendall(1938)}]{kendall1938new}
Maurice~G Kendall. 1938.
\newblock A new measure of rank correlation.
\newblock \emph{Biometrika}, 30(1-2):81--93.

\bibitem[{Krosnick(1991)}]{krosnick1991response}
Jon~A Krosnick. 1991.
\newblock Response strategies for coping with the cognitive demands of attitude measures in surveys.
\newblock \emph{Applied cognitive psychology}, 5(3):213--236.

\bibitem[{Lin(2004)}]{ROUGE2004}
Chin-Yew Lin. 2004.
\newblock \href {https://aclanthology.org/W04-1013/} {{ROUGE}: A package for automatic evaluation of summaries}.
\newblock In \emph{Text Summarization Branches Out}, pages 74--81. Association for Computational Linguistics.

\bibitem[{Liu et~al.(2023)Liu, Iter, Xu, Wang, Xu, and Zhu}]{GEval2023}
Yang Liu, Dan Iter, Yichong Xu, Shuohang Wang, Ruochen Xu, and Chenguang Zhu. 2023.
\newblock \href {https://arxiv.org/abs/2303.16634} {G-eval: Nlg evaluation using gpt-4 with better human alignment}.
\newblock \emph{CoRR}.

\bibitem[{Liu et~al.(2024)Liu, Yang, Huang, Zhang, Huang, Wei, Deng, Sun, and Zhang}]{HD-Eval2024}
Yuxuan Liu, Tianchi Yang, Shaohan Huang, Zihan Zhang, Haizhen Huang, Furu Wei, Weiwei Deng, Feng Sun, and Qi~Zhang. 2024.
\newblock \href {https://arxiv.org/abs/2402.15754} {Hd-eval: Aligning large language model evaluators through hierarchical criteria decomposition}.
\newblock \emph{CoRR}.

\bibitem[{Netzer et~al.(2012)Netzer, Feldman, Goldenberg, and Fresko}]{netzer2012mine}
Oded Netzer, Ronen Feldman, Jacob Goldenberg, and Moshe Fresko. 2012.
\newblock Mine your own business: Market-structure surveillance through text mining.
\newblock \emph{Marketing Science}, 31(3):521--543.

\bibitem[{Papineni et~al.(2022)Papineni, Roukos, Ward, and Zhu}]{BLEU2002}
Kishore Papineni, Salim Roukos, Todd Ward, and Wei-Jing Zhu. 2022.
\newblock \href {https://doi.org/10.3115/1073083.1073135} {{B}leu: a method for automatic evaluation of machine translation}.
\newblock In \emph{ACL}, pages 311--318.

\bibitem[{Park and Cho(2014)}]{konlpy@2014}
Eunjeong~L. Park and Sungzoon Cho. 2014.
\newblock Konlpy: Korean natural language processing in python.
\newblock In \emph{HCLT}.

\bibitem[{Reimers and Gurevych(2019)}]{SBERT2019}
Nils Reimers and Iryna Gurevych. 2019.
\newblock Sentence-{BERT}: Sentence embeddings using {S}iamese {BERT}-networks.
\newblock In \emph{EMNLP-IJCNLP}, pages 3982--3992.

\bibitem[{Rouder et~al.(2021)Rouder, Saucier, Kinder, and Jans}]{Rouder2021What}
Jessie Rouder, Olivia Saucier, Rachel Kinder, and Matt Jans. 2021.
\newblock \href {https://doi.org/10.29115/SP-2021-0008} {What to {Do} {With} {All} {Those} {Open}-{Ended} {Responses}? {Data} {Visualization} {Techniques} for {Survey} {Researchers}}.
\newblock \emph{Survey Practice}.

\bibitem[{Santilli et~al.(2025)Santilli, Golinski, Kirchhof, Danieli, Blaas, Xiong, Zappella, and Williamson}]{santilli2025revisitinguncertaintyquantificationevaluation}
Andrea Santilli, Adam Golinski, Michael Kirchhof, Federico Danieli, Arno Blaas, Miao Xiong, Luca Zappella, and Sinead Williamson. 2025.
\newblock \href {https://arxiv.org/abs/2504.13677} {Revisiting uncertainty quantification evaluation in language models: Spurious interactions with response length bias results}.
\newblock \emph{CoRR}.

\bibitem[{Spearman(1987)}]{spearman1987proof}
Charles Spearman. 1987.
\newblock The proof and measurement of association between two things.
\newblock \emph{The American journal of psychology}, 100(3/4):441--471.

\bibitem[{Wu et~al.(2025)Wu, Hossain, Wood, Akbar, Chin, and Cornejo}]{SeeEval2025}
Meng-Chen Wu, Md~Mosharaf Hossain, Tess Wood, Shayan~Ali Akbar, Si-Chi Chin, and Erwin Cornejo. 2025.
\newblock \href {https://aclanthology.org/2025.findings-naacl.411/} {{SEE}val: Advancing {LLM} text evaluation efficiency and accuracy through self-explanation prompting}.
\newblock In \emph{NAACL}, pages 7357--7368.

\bibitem[{Zhang et~al.(2020)Zhang, Kishore, Wu, Weinberger, and Artzi}]{BERTScore2020}
Tianyi Zhang, Varsha Kishore, Felix Wu, Kilian~Q. Weinberger, and Yoav Artzi. 2020.
\newblock \href {https://openreview.net/forum?id=SkeHuCVFDr} {Bertscore: Evaluating text generation with {BERT}}.
\newblock In \emph{ICLR}.

\bibitem[{Zheng et~al.(2023)Zheng, Chiang, Sheng, Zhuang, Wu, Zhuang, Lin, Li, Li, Xing, Zhang, Gonzalez, and Stoica}]{LLMasJudge2023}
Lianmin Zheng, Wei{-}Lin Chiang, Ying Sheng, Siyuan Zhuang, Zhanghao Wu, Yonghao Zhuang, Zi~Lin, Zhuohan Li, Dacheng Li, Eric~P. Xing, Hao Zhang, Joseph~E. Gonzalez, and Ion Stoica. 2023.
\newblock \href {http://papers.nips.cc/paper\_files/paper/2023/hash/91f18a1287b398d378ef22505bf41832-Abstract-Datasets\_and\_Benchmarks.html} {Judging llm-as-a-judge with mt-bench and chatbot arena}.
\newblock In \emph{NeurIPS}.

\end{thebibliography}


\newpage
\clearpage
\appendix

\section{Language-Specific Gibberish Detection}
\label{appendix:gibberish}

We designed a gibberish detection module to identify nonsensical or meaningless text. This module helps reduce evaluation costs by filtering out samples that can be automatically assessed prior to human evaluation. The module consists of three stages: (i) \textbf{Preprocessing}, (ii) \textbf{Markov modeling}, and (iii) \textbf{Language-Specific Filtering}. A sentence is classified as gibberish if either its average log-likelihood falls below a certain threshold(\eg, -4.0, -3.5) or it fails any of the language-specific filters described above. Since the characteristics and generative patterns of gibberish vary significantly across languages, detection strategies must be tailored to the structural and statistical properties of each language. In this study, we designed two independent detection pipelines, one for English and one for Korean, each reflecting the unique linguistic characteristics of the respective language.

\subsection{Characteristics and Detection Strategy for English Gibberish}

\subsubsection{Preprocessing}
The input text is converted to lowercase, retaining only Latin letters(a–z) and whitespace. Punctuation, digits, and other non-alphabetic symbols are removed to reduce noise and focus on purely alphabetic content.

\subsubsection{Markov Model}
\label{sec:app_markov}
We constructed a character-level bigram Markov model based on real-world conversational data\footnote{\url{https://www.kaggle.com/datasets/projjal1/human-conversation-training-data}}. Using this model, the average log-likelihood of a given input sentence is computed, providing a statistical measure of how well the text conforms to typical English character sequences. The threshold for the Markov model was empirically determined to be -4.0.

\subsubsection{Language-Specific Filtering}

\begin{itemize}
    \item \textbf{Consecutive Consonant/Vowel Sequence Detection:} We detect sequences in which vowels or consonants are unusually repeated (e.g., \texttt{aaaaaa}, \texttt{bbbbb}). Such patterns often arise from keyboard mashing or unintended input. If the length of any such sequences exceeds a predefined threshold(\, 10), the text is classified as gibberish.
    \item \textbf{Valid Word Ratio:} Using the NLTK tokenizer~\cite{  nltk@2009} and English dictionary, we calculate the proportion of valid English words in each response. If this ratio falls below a predefined threshold (\eg, 0.4), the text is classified as gibberish.
    \item \textbf{Exception Handling for Abbreviations and Slang:}
    English conversational text frequently contains non-standard tokens (\eg, abbreviations, slang, or elongated forms) that may superficially resemble gibberish. We retain such tokens when they convey clear meaning, or when either (i) the average log-likelihood under the English bigram Markov model exceeds $-4.0$ or (ii) the valid-word ratio surpasses the threshold. Frequent items (e.g., \emph{lol}, \emph{brb}) and laughter variants (e.g., \emph{ha}, \emph{haha}) are kept; elongated forms (e.g., \emph{soooo}) are also retained unless both criteria fail and the text exhibits repetitive, low-information patterns. Purely nonsensical sequences (e.g., \emph{asdf}, \emph{dddd}) remain filtered out.
\end{itemize}

\subsection{Characteristics and Detection Strategy for Korean Gibberish}

\subsubsection{Preprocessing}
Korean text is written in syllabic blocks, each composed of individual Jamo characters (i.e., consonants and vowels). To capture fine-grained statistical patterns, we first decompose each syllable block into its constituent Jamo units.

\subsubsection{Markov Model}
Following decomposition, we designed a Jamo-level bigram Markov model based on the frequency of Jamo character pairs (i.e., sequences of initial consonant, medial vowel, and final consonant) of real-world conversational data\footnote{\url{https://github.com/Ludobico/KakaoChatData}}. This model estimates the average log-likelihood of a given text sequence, reflecting how well the character transitions conform to typical Korean orthographic patterns. The threshold for the Markov model was empirically set to -3.5.

\subsubsection{Language-Specific Filtering}
\begin{itemize}
    \item \textbf{Syllabic Completeness and Jamo Diversity:} The proportion of fully-formed Hangul syllables and the diversity of unique Jamo components are used as additional indicators to capture repetitive or structurally abnormal sequences. If the proportion of complete Hangul syllables falls below a specified threshold (e.g., 0.6), or if the diversity of unique Jamo characters is below a certain value (e.g., 4), the input is classified as gibberish.
    \item \textbf{Morphological Analysis Filtering:} We apply morphological analysis using KoNLPy's \texttt{Okt} tokenizer~\cite{konlpy@2014}. Sentences that fail to produce any recognizable morphemes are directly labeled as gibberish, as they likely contain no valid lexical units.
    \item  \textbf{Exception Handling for Neologisms:} Certain forms of non-standard expressions in Korean-such as neologisms, abbreviations, or repetitive emotional markers may resemble gibberish due to their unconventional structure or brevity. Therefore, we manually whitelisted common conversational responses such as “ㅋㅋㅋ”, “ㅇㅇ”, or “헐” to prevent misclassification, despite their repetitive characters or absence from formal lexicons.
\end{itemize}

\subsection{Experiment}
\subsubsection{Baseline}
\paragraph{Pretrained Gibberish Detector}
We employed the publicly available pretrained model~\cite{gibberishdetector@2021} from Hugging Face. This model is a BERT-based classifier trained specifically to detect gibberish in English text.

\paragraph{Markov Model}
We employed a character-level bigram-based Markov probability model. The model collects the frequencies of adjacent character or jamo pairs from a training corpus and computes the average log-likelihood for a given input text. If the resulting value falls below a predefined threshold, the input is classified as gibberish. This approach corresponds to our method without the language-specific filtering step.

\paragraph{LLM}
Next, we adopted a prompting approach using a LLM (\texttt{gpt-4o-mini-2024-07-18}).
The model was instructed to output either “true” or “false,” indicating whether the given sentence is meaningful (i.e., not gibberish).

\begin{figure*}[t]
\centering
\includegraphics[width=1.0\linewidth]{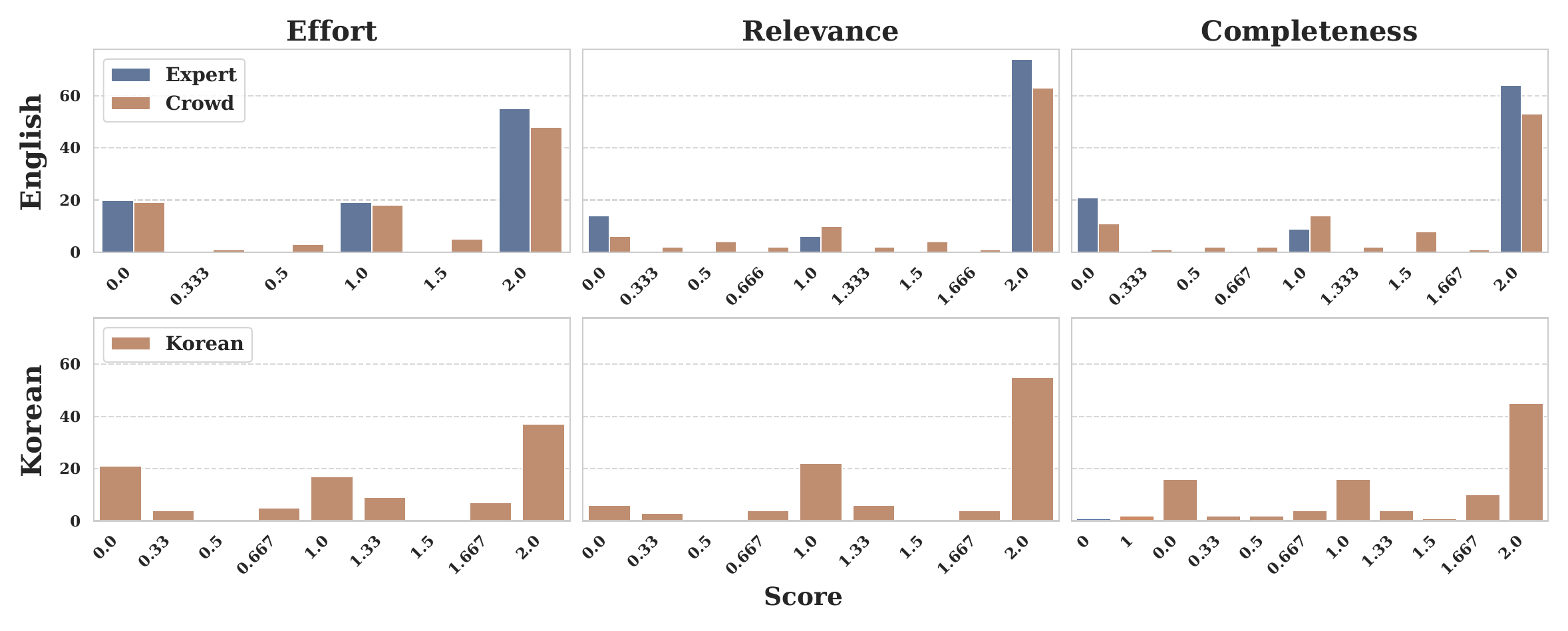} 
\caption{Score distribution of annotations by expert and crowd workers}
\label{fig:score_dist}
\end{figure*}

\subsubsection{Dataset}
For the test datasets, we collected 100 real-world samples for both English and Korean, each containing 50 gibberish and 50 non-gibberish sentences.

\subsubsection{Results}
\begin{table}[ht]
\centering
\small
\resizebox{\columnwidth}{!}{%
\begin{tabular}{llcccc}
\toprule
\textbf{Lang.} & \textbf{Model} & \textbf{F1} & \textbf{Prec.} & \textbf{Recall} & \textbf{Acc.} \\
\midrule
\multirow{4}{*}{En} 
  & BERT & 0.8900 & 0.9100 & 0.8900 & 0.8900 \\
  & Markov      & 0.3967 & 0.7577 & 0.5300 & 0.5300 \\
  & LLM         & 0.9247 & \textbf{1.0000} & 0.8600 & 0.9300 \\
  & \textbf{Ours}        & \textbf{0.9800} & 0.9810 & \textbf{0.9800} & \textbf{0.9800} \\
\midrule
\multirow{3}{*}{Ko} 
  & Markov      & 0.6011 & 0.7941 & 0.6500 & 0.6500 \\
  & LLM         & 0.9474 & \textbf{1.0000} & 0.9000 & 0.9500 \\
  & \textbf{Ours}        & \textbf{0.9800} & 0.9810 & \textbf{0.9800} & \textbf{0.9800} \\
\bottomrule
\end{tabular}
}
\caption{Gibberish detection performance across languages. The table reports F1 score, Prec.(Precision), Recall, and Acc.(Accuracy) for English (En) and Korean (Ko) datasets using different models.}
\label{tab:gibberish_results}
\end{table}

Table~\ref{tab:gibberish_results} presents the gibberish detection performance across baseline models on both English and Korean datasets. Four metrics were used for evaluation: F1 score, Precision, Recall, and Accuracy.

\paragraph{English Dataset}
On the English dataset, the pretrained gibberish detector provided by Hugging Face achieved reasonable performance, with an F1 score and accuracy both at 0.89. The LLM performed even better, yielding an F1 score of 0.9247 and an accuracy of 0.93. In contrast, the character-level Markov model showed significantly lower performance, with an F1 score of 0.3967 and a recall of 0.53, indicating that this method failed to identify many gibberish instacnes.  
Our proposed model outperformed all baselines, achieving the best performance across all evaluation metrics, with F1, precision, recall, and accuracy all reaching 0.98.

\paragraph{Korean Dataset}
On the Korean dataset, overall performance across models was slightly higher than on the English dataset. This improvement can be attributed to the structural characteristics of Korean, where Jamo-level analysis enables more effective detection of gibberish.
The Markov model outperformed its English counterpart, achieving an F1 score of 0.6011 and an accuracy of 0.65, though it still struggled with recall.  
The LLM demonstrated perfect precision (1.0) and a strong F1 score of 0.9474, but slightly lower recall (0.9), missing a few gibberish instances.  
Our proposed model once again achieved the best performance across all evaluation metrics, matching its English results with F1, precision, recall, and accuracy all at 0.98.

\paragraph{Limitations and Comparative Findings}
The Markov model, while computationally simple and fast, suffers from the limitation of relying solely on character-level frequency patterns, making it ineffective at detecting more complex or irregular gibberish.  
On the other hand, both the pretrained gibberish detector and LLM attained relatively high precision but exhibited limitations in consistency and cross-lingual adaptability.  
Ultimately, our proposed method, which integrates structural characteristics and auxiliary linguistic signals (e.g., Jamo diversity, syllable completeness, and morphological analysis), achieved the most robust and reliable results across both languages.

\section{Dataset and Annotation Details}
\label{appendix:dataset}

\subsection{Dataset Overview}

In this section, we provide details on the datasets and annotation procedures. The responses were written by actual survey participants and were sampled to ensure diversity in response lengths and quality levels. The English dataset consists of 275 responses (181 for dev and 94 for test), and the Korean dataset consists of 222 responses (122 for dev and 100 for test).

\subsection{Annotation Procedure}
\subsubsection{Annotation Workflow}
\paragraph{English Dataset}
All responses were evaluated by either crowd-workers or domain experts.
For the English dev set, 1–3 crowd-workers from Prolific\footnote{\url{https://www.prolific.com/}} annotated each response along four dimensions: \textit{effort}, \textit{relevance}, \textit{completeness}, and gibberish.
In the case of the English test set, a single domain expert annotated all responses across all dimensions. 

\paragraph{Korean Dataset}
For the Korean dataset, responses were translated into English before being annotated by the domain expert. The Korean-to-English translation process consisted of two stages. 
In the first stage, we employed ChatGPT to generate an initial draft 
translation. To preserve the authentic characteristics of the real-world human responses, the model specifically prompted to perform a direct translation that prioritized fidelity to the original text over fluency, thereby retaining grammatical errors, typographical mistakes, and unconventional sentence structures. 
In the second stage, an internal bilingual reviewer examined each translated sentence against its original Korean counterpart to correct inaccuracies and ensure that subtle nuances, including the original errors or stylistic variations potentially overlooked by the LLM, were faithfully represented.

Finally, expert annotations were collected for the overall quality and rejection dimensions of the English and Korean test sets to ensure alignment with marketing research practice. Figure~\ref{fig:score_dist} shows the score distributions of experts and crowd workers for the English and Korean datasets.

\subsubsection{Human Annotation Guidelines}
\label{appendix:human_annotation}
Responses were sampled from prior user studies to ensure diversity in question types, response lengths, and quality levels. Each sample was fully annotated across all four evaluation dimensions: \textit{effort}, \textit{relevance}, \textit{completeness}, and \textit{overall quality}. Each response was annotated by 1 to 3 raters using a Likert scale. In addition, acceptance was categorized into three distinct classes.

\begin{itemize}[leftmargin=5mm]
\item \textbf{Effort} (0–2): Low (0), Okay (1), Good (2)
\item \textbf{Relevance} (0–2): Irrelevant (0), Partially Relevant (1), Relevant (2)
\item \textbf{Completeness} (0–2): Not Fulfilled (0), Partially Fulfilled (1), Fully Fulfilled (2)
\item \textbf{Overall quality} (0–4): Very Poor (0), Acceptable (1), Fair (2), Good (3), Excellent (4)
\item \textbf{Acceptance}: Accept, Hold, Reject
\end{itemize}


\section{Further Analyses}
\label{appendix:correlation_analysis}
\subsection{Correlation Analysis}

Table~\ref{tab:correlation} presents the correlation results between expert annotations and crowd-sourced annotations for each evaluation dimension—\textit{effort}, \textit{relevance}, \textit{completeness}—as well as \textit{overall quality}. Specifically, it reports the correlations coefficients among the following pairs: Expert–Ours, Crowd–Ours, and Expert–Crowd, along with inter-annotator agreement among the crowd annotators.

As shown in the results tables, both Expert-Ours and Crowd-Ours correlations are high for English, with Expert-Ours correlations being higher. This indicates that our proposed method aligns more closely with expert evaluations for English datasets.

For Korean, however, the correlation with experts is lower in overall quality compared to that with the crowd. This may be due to the fact that expert annotations were based on English translations of Korean responses, which could have distorted subtle linguistic nuances and contextual cues.

\subsection{Significance Test for Correlation}
\paragraph{Methodology}
We estimated bootstrap confidence intervals (CIs) for the differences in correlation coefficients between our proposed method and the second-best baseline (i.e., the method with the next-highest correlation) across the three dimensions \textit{effort}, \textit{relevance}, and \textit{completeness}, and across two languages, English and Korean. Specifically, we performed bootstrap resampling with replacement at the level of question–response pairs, repeatedly computing the differences in Spearman’s ρ and Kendall’s τ. The final 95\% CIs were derived using the percentile method. If the CI did not include 0, the result was considered statistically significant.

\begin{figure*}[t]
\centering
\includegraphics[width=1.0\linewidth]{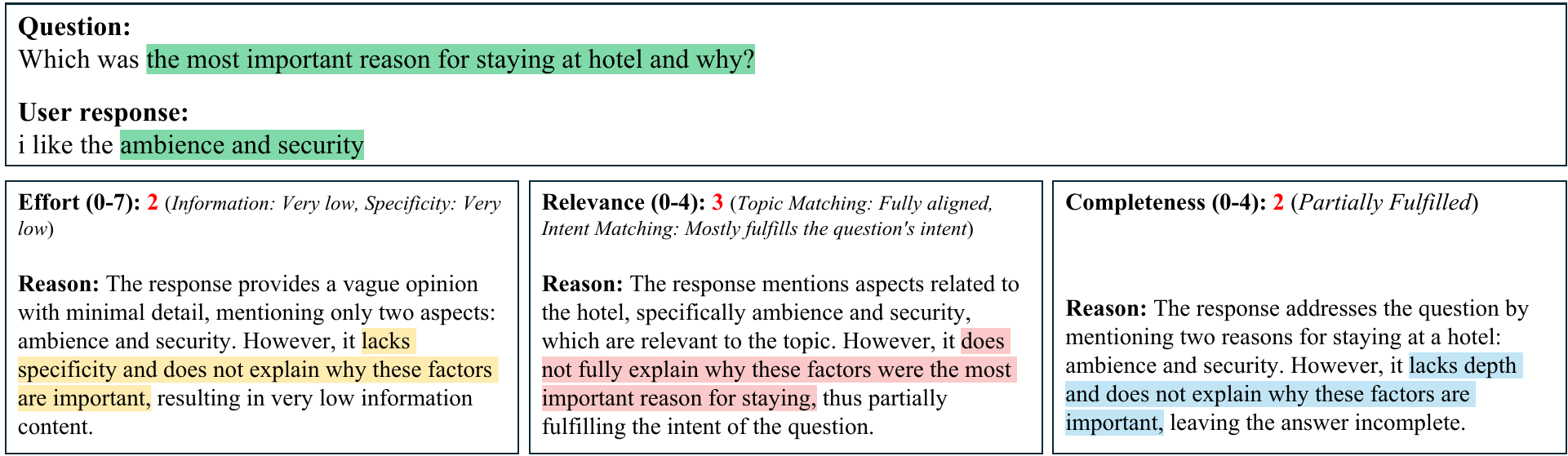} 
\caption{A case study presenting the evaluation results of a single user response using our method in English dataset.}
\label{fig:casestudy_en}
\end{figure*}
\begin{figure*}[t]
\centering
\includegraphics[width=1.0\linewidth]{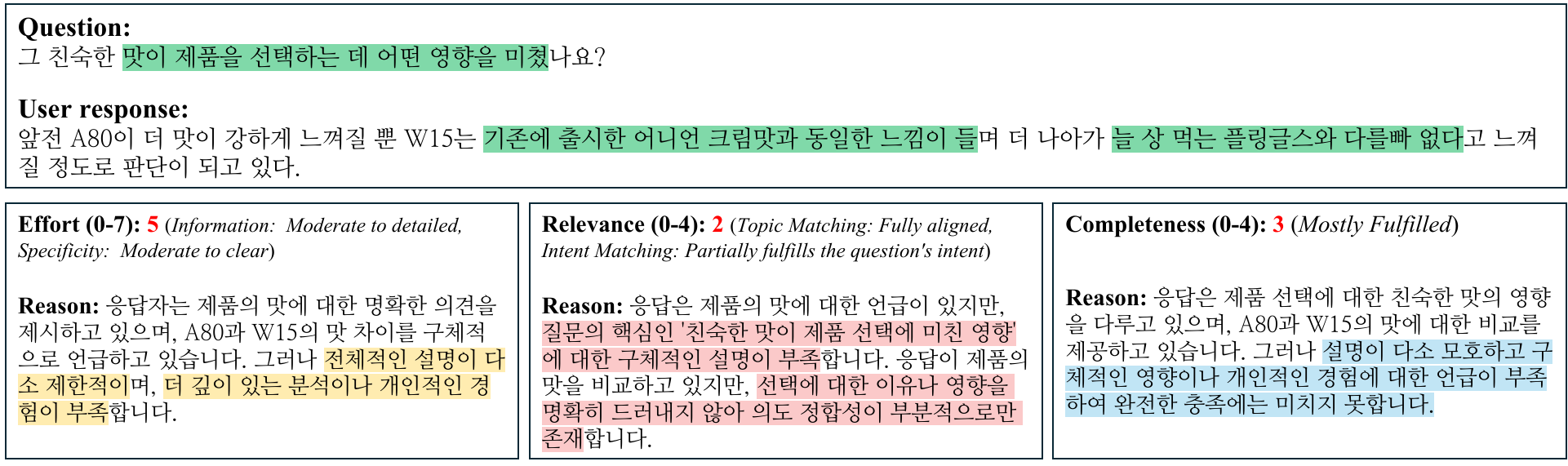} 
\caption{A case study presenting the evaluation results of a single user response using our method in Korean dataset.}
\label{fig:casestudy_ko}
\end{figure*}

\paragraph{Results and Analysis}
\begin{table*}[t]
\small
\centering
\begin{tabular}{lcccc}
\toprule
& \multicolumn{2}{c}{\textbf{English}} & \multicolumn{2}{c}{\textbf{Korean}} \\
\cmidrule(lr){2-3}\cmidrule(lr){4-5}
\textbf{Dimension} & \textbf{Spearman $\rho$} & \textbf{Kendall $\tau$} & \textbf{Spearman $\rho$} & \textbf{Kendall $\tau$} \\
\midrule
Effort       & $[-0.005,\, 0.075]$  & $[0.033,\, 0.124]$ & $[-0.048,\, 0.143]$  & $[-0.023,\, 0.160]$ \\
Relevance    & $[0.079,\, 0.329]$   & $[0.075,\, 0.301]$ & $[-0.082,\, 0.195]$ & $[-0.017,\, 0.227]$ \\
Completeness & $[-0.010,\, 0.087]$ & $[0.044,\, 0.153]$ & $[-0.003,\, 0.211]$ & $[0.029,\, 0.229]$ \\
\bottomrule
\end{tabular}
\caption{Bootstrap 95\% confidence intervals (CIs) of correlation differences (Ours vs. second-best baseline).}
\label{tab:tab_CIs}
\end{table*}

Table~\ref{tab:tab_CIs} presents the 95\% bootstrap confidence intervals for the correlation differences between our method and the second-best baseline across dimensions and languages.

\begin{itemize}[leftmargin=5mm]
\item \textbf{Significant improvements:} We observe statistically significant improvements in English \textit{relevance} (Spearman and Kendall) and Korean \textit{completeness} (Kendall), where the CIs exclude 0. This indicates that our method is particularly effective at capturing semantic alignment (\textit{relevance}) and core informational fulfillment (\textit{completeness}).
    
\item \textbf{Effort:} Since \textit{effort} is inherently correlated with superficial features such as text length (with the second-best baseline being token-length), differences were not statistically significant in most cases. Nevertheless, English Kendall consistently shows a positive and significant improvement, highlighting advantages in rank-order consistency.
    
\item \textbf{Impact of sample size:} The relatively small test set sizes (English: 94, Korean: 100) led to wide confidence intervals. We expect that validation on larger real-world datasets will yield clearer patterns of statistical significance across all dimensions.
\end{itemize}

\begin{table*}[]
\small
\centering
\begin{tabular}{cc|cc|cc}
\toprule
\multicolumn{2}{c|}{\textbf{Language}}                   & \multicolumn{2}{c|}{\textbf{English}}    & \multicolumn{2}{c}{\textbf{Korean}}    \\ \cmidrule{3-6} 
\textbf{Dimension}                     & \textbf{Method} & \textbf{Spearman $\rho$} & \textbf{Kendall $\tau$} & \textbf{Spearman $\rho$} & \textbf{Kendall $\tau$} \\ \hline
\multirow{4}{*}{Effort}       & Expert vs. Ours         & 0.8198 & 0.7223 & -      & -      \\
                              & Crowd vs. Ours          & 0.8086 & 0.7155 & 0.7950 & 0.6797 \\
                              & Expert vs. Crowd        & 0.7036 & 0.6489 & -      & -      \\
                              & Inter-annotator (crowd) & 0.7986 & 0.7089 & 0.7385 & 0.6867 \\ \midrule
\multirow{4}{*}{Relevance}    & Expert vs. Ours         & 0.8614 & 0.7954 & -      & -      \\
                              & Crowd vs. Ours          & 0.6575 & 0.5822 & 0.6021 & 0.5269 \\
                              & Expert vs. Crowd        & 0.6610 & 0.6087 & -      & -      \\
                              & Inter-annotator (crowd) & 0.7940 & 0.7288 & 0.6784 & 0.6432 \\ \midrule
\multirow{4}{*}{Completeness} & Expert vs. Ours         & 0.8245 & 0.7438 & -      & -      \\
                              & Crowd vs. Ours          & 0.7374 & 0.6532 & 0.7457 & 0.6372 \\
                              & Expert vs. Crowd        & 0.6438 & 0.5890 & -      & -      \\
                              & Inter-annotator (crowd) & 0.7799 & 0.7029 & 0.6007 & 0.5676 \\ \midrule
\multirow{4}{*}{Overall quality (Sum)} & Expert vs. Ours & 0.8953              & 0.7820             & 0.6024              & 0.5066           \\
                              & Crowd vs. Ours          & 0.8318 & 0.7003 & 0.7040 & 0.5478 \\
                              & Expert vs. Crowd        & 0.7792 & 0.6822 & 0.5318 & 0.4559 \\
                              & Inter-annotator (crowd) & 0.8654 & 0.7517 & 0.6127 & 0.5034 \\ \bottomrule
\end{tabular}
\caption{Correlations between Expert and Ours, Crowd and Ours, and Expert and Crowd. Crowd refers to annotators recruited via Prolific. Inter-annotator indicates the agreement among crowd annotators.}
\label{tab:correlation}
\end{table*}

\subsection{Quadratic-weighted Kappa}
\begin{table}[]
\small
\centering
\begin{tabular}{l|ll}
\toprule
\textbf{Dimension}             & \textbf{English} & \textbf{Korean} \\ \midrule
Effort       & 0.3632  & 0.4001 \\
Relevance    & 0.4463  & 0.4227 \\
Completeness & 0.6242  & 0.5864 \\ \bottomrule
\end{tabular}

\caption{Performance of the proposed method and human agreement using Quadratic-weighted Kappa.}
\label{tab:kappa}
\end{table}
Table~\ref{tab:kappa} shows the inter-rater agreement between the proposed evaluator and human raters, measured using the quadratic-weighted kappa.
Notably, the \textit{completeness} dimension exhibits the highest importance across both languages,
suggesting that the LLM-based evaluator applying our analytically defined evaluation criteria can achieve a high level of agreement with human judgments.





\subsection{Length-based Baselines for Effort}
\begin{table}[t]
\centering
\small
\begin{tabular}{lc}
\toprule
\textbf{\# Sentences} & \textbf{Count} \\
\midrule
1  & 57 \\
2  & 19 \\
3  & 9  \\
4  & 6  \\
8  & 1  \\
9  & 1  \\
30 & 1  \\
\bottomrule
\end{tabular}
\caption{Distribution of responses by number of sentences (English test set).}
\label{tab:g_sentence_dist}
\end{table}

\begin{table}[t]
\centering
\small
\begin{tabular}{lcc}
\toprule
\textbf{Length-based proxy} & \textbf{Spearman $\rho$} & \textbf{Kendall $\tau$} \\
\midrule
Sentence count-based         & $-0.0616$ & $-0.0564$ \\
Noun/Adj./Verb count-based   & $-0.0629$ & $-0.0569$ \\
\bottomrule
\end{tabular}
\caption{Correlation with expert Effort ratings using alternative length-based proxies (English test set).}
\label{tab:g_corr_lenvariants}
\end{table}

We also evaluated alternative length-based baselines beyond token counts, including sentence counts (\textbf{Sentence count-based}) and counts of nouns, adjectives, and verbs (excluding \emph{be}-verbs) (\textbf{Noun/Adj./Verb count-based}).\footnote{For part-of-speech extraction, we used the NLTK toolkit \cite{  nltk@2009}. Sentence counts were computed by splitting on punctuation marks, primarily periods.}

As shown in Table~\ref{tab:g_sentence_dist}, human responses are generally short and informal, with most consisting of only one or two sentences. Many are written as a single run-on sentence or phrase. Because of these characteristics, Sentence count-based and Noun/Adj./Verb count-based measures failed to capture meaningful variation in response length, leading to considerably weaker correlations with expert \textit{effort} ratings.

Table~\ref{tab:g_corr_lenvariants} reports Spearman’s $\rho$ and Kendall’s $\tau$ when \textit{effort} is approximated by sentence counts and by counts of nouns, adjectives, and verbs (excluding \emph{be}-verbs). As shown, these correlations are substantially lower than those based on token length. Given the short and highly informal nature of the responses, we therefore adopt token count as the length baseline in the main analysis.

\section{Case Study}
\label{appendix:case_study}
We conducted a real-world case study of English and Korean data samples to validate the interpretability and effectiveness of our proposed evaluation framework.
\paragraph{English}
In the English example shown in Figure~\ref{fig:casestudy_en}, the user answers "i like the ambience and security" to the question "Which was the most important reason for staying at the hotel and why?". The response receives low scores for both \textit{effort} (2/7) and \textit{completeness} (2/4), as it merely lists two factors-ambience and security-without elaboration or justification. However, the \textit{relevance} score is relatively higher (3/4), since the mentioned aspects are directly related to the topic of hotel preference and partially fulfill the intent of the question.
\paragraph{Korean}
In the Korean example shown in Figure~\ref{fig:casestudy_ko}, the users answer the question "How did that familiar flavor influence your choice of product?" by comparing the flavors of two snack products (A80 and W15), noting that W15 tastes identical to an existing onion cream flavor and even resembles a familiar brand like Pringles. This response scores relatively high in \textit{effort} (5/7) and \textit{completeness} (3/4) due to its concrete product comparisons and moderate cognitive engagement. However, its \textit{relevance} is rated lower (2/4) because, while it discusses taste, it does not directly address the core question, which is about how the familiarity of the flavor influenced the respondent’s product choice.

This discrepancy highlights the importance of evaluating multiple distinct dimensions of response quality. Relying on a single aspect could obscure critical weaknesses or strengths. Moreover, the dimension-specific explanations provided by the framework enhance interpretability, offering valuable insights into the nature of each response.

\section{LLM Evaluation Prompt}\label{appendix:prompts}

This section presents the LLM evaluation prompts used for the four assessment dimensions—\textit{effort}, \textit{relevance}, \textit{completeness}, and \textit{overall quality}—as introduced in Section~\ref{dimension_eval_method} and Section~\ref{overall_quality_method}. Each dimensional prompt is tailored to a specific dimension and includes a defined scoring range, evaluation criteria, and illustrative response examples. The prompts are designed based on rubrics to support consistent and reliable judgments by the evaluator. 

The \textit{effort} dimension assesses the respondent’s cognitive engagement, focusing on the amount of information provided and the specificity of the response. The \textit{relevance} dimension evaluates how well the response aligns with both the topic and the intent of the question. The \textit{completeness} dimension measures the extent to which the response addresses the core informational requirements of the question. The \textit{overall quality} dimension integrates the scores and justifications from the three preceding dimensions to produce a comprehensive assessment of response quality.

The LLM evaluates each dimension according to the corresponding prompt and returns both a discrete score and a brief justification. This prompt-based evaluation approach contributes to greater transparency, consistency, and interpretability in scoring.

The prompts were initially developed in English and then directly translated for use with the Korean data. Detailed English prompt specifications are provided in Tables~\ref{tab:prompt_effort_v4}--~\ref{tab:prompt_overall_quality}.

\begin{table*}[ht] \small
\centering
\begin{tabular}{p{16cm}}
\specialrule{1pt}{0pt}{1pt}
\textbf{System Prompt:} \\
You are an expert evaluator for a human-AI conversation dataset. Your task is to evaluate the quality of the responses in the dataset based on the given criteria. \\
\specialrule{0.4pt}{1pt}{1pt}
\textbf{User Prompt:} \\
Rate how much thought and detail the user put into the response. \\
Use the following \textbf{0–7 scale}. The score must be an integer. \\
Base your judgment on the \textbf{information content}, \textbf{specificity}, and \textbf{how well the user responds to the question}. \\ \\
Examples are provided to guide interpretation: \\[1ex]

\textbf{0}: No meaningful response. The answer is either empty or completely unrelated. \\
\quad • Information: None \\
\quad • Specificity: None \\
\quad • Response to question: Not at all \\
\quad e.g., “N/A”, “blah”, “asdf” \\[1ex]

\textbf{1}: Vague or evasive, such as default or placeholder answers that avoid the question. \\
\quad • Information: Minimal or token or negligible \\
\quad • Specificity: None \\
\quad • Response to question: Barely reacts, without offering any insight or detail \\
\quad e.g., “Good”, “Okay”, “I don't know”, “Maybe” \\[1ex]

\textbf{2}: Vague or generic opinion. Slightly more than a one-word answer, but still lacking substance. \\
\quad • Information: Very low \\
\quad • Specificity: Very low \\
\quad • Response to question: Barely reacts \\
\quad e.g., “Pretty good overall”, “Not bad”, “Nice” \\[1ex]

\textbf{3}: A short opinion that includes a single element or impression. \\
\quad • Information: Low \\
\quad • Specificity: Low \\
\quad • Response to question: Partially addresses one aspect \\
\quad e.g., “Liked the burger”, “Sick fries”, “Nice vibe” \\[1ex]

\textbf{4}: Slightly more informative with two aspects mentioned, but still minimal explanation. \\
\quad • Information: Slightly basic \\
\quad • Specificity: Limited \\
\quad • Response to question: Briefly addresses two parts \\
\quad e.g., “Burger was good. Service okay”, “Decent food but small portions” \\[1ex]

\textbf{5}: Thoughtful response with several clear opinions and specific examples. \\
\quad • Information: Moderate to detailed \\
\quad • Specificity: Moderate to clear \\
\quad • Response to question: Engages with key parts \\
\quad e.g., “Tasty food, but nothing special.”, “Fries were crisp and burger was hot, but too salty.” \\[1ex]

\textbf{6}: Very rich and nuanced response; explains what stood out and why. \\
\quad • Information: Very high \\
\quad • Specificity: Very high \\
\quad • Response to question: Deeply addresses the question with insight \\
\hangindent=1em
\hangafter=0
e.g., “The burger had a smoky flavor, fries were hot and crisp, and the vintage vibe was cool. Pricey, but worth it.” \\[1ex]

\textbf{7}: Exceptionally detailed, thoughtful, and complete. Evaluates multiple dimensions (e.g., food, service, atmosphere) with depth and personality. \\
\quad • Information: Comprehensive \\
\quad • Specificity: Deep \\
\quad • Response to question: Fully and thoughtfully addressed \\
\hangindent=1em
\hangafter=0
e.g., “Burger exceeded expectations. The double bacon burger was perfectly cooked, the bun was fresh, fries had just the right crunch. Staff were welcoming, and the drive-in movie setup made the experience unique.” \\[1ex]

Read the given criteria carefully and follow them faithfully. \\
Return your evaluation as a JSON dictionary without any additional text: \\[1ex]

\texttt{\{} \\
\texttt{\quad "effort": <int: 0-7>,} \\
\texttt{\quad "reason": "Detailed justification based on the criteria (2–3 sentences)"} \\
\texttt{\}} \\[1ex]

Now, let’s get it started! \\
\texttt{Conversation: \{conversation\}} \\
\texttt{Your evaluation: \{JSON dictionary\}} \\
\specialrule{1pt}{0pt}{1pt}
\end{tabular}
\caption{Prompt used for \textit{effort} evaluation.}
\label{tab:prompt_effort_v4}
\end{table*}



\begin{table*}[ht] \small
\centering
\begin{tabular}{p{16cm}}
\specialrule{1pt}{0pt}{1pt}
\textbf{System Prompt:} \\
You are an expert evaluator for a human-AI conversation dataset. Your task is to evaluate the quality of the responses in the dataset based on the given criteria. \\
\specialrule{0.4pt}{1pt}{1pt}
\textbf{User Prompt:} \\
Rate how well the response aligns with the topic and intent of the question. \\[1ex]
\textbf{Two criteria} should be considered: \\
- \textbf{Topic alignment}: Is the response about the same subject as the question? \\
- \textbf{Intent alignment}: Does the response address the purpose behind the question? \\[1ex]
\textbf{Note}: The level of detail, reasoning, or length of the response should not be considered here. Even a short or simple response can receive a high relevance score if it correctly addresses the question's intent. \\ \\ 
Examples are provided to guide interpretation: \\[1ex]
\textbf{0: Completely irrelevant or non-responsive} \\
The response is off-topic, nonsensical, or fails to engage with the question in any meaningful way. \\
\quad • Topic alignment: No \\
\quad • Intent alignment: No \\
\hangindent=1em
\hangafter=0
Example: Q: "Can you describe your experience with this product?", A: "I don't know." / "Get rich." \\[1ex]
\textbf{1: Topic is mentioned, but intent completely missed} \\
The response includes a term or concept related to the question but does not address the question’s actual purpose. \\
\quad • Topic alignment: Yes \\
\quad • Intent alignment: No \\
\hangindent=1em
\hangafter=0
Example: Q: "Why did you choose this brand?", A: "The logo looks nice." (Mentions the topic, but doesn’t explain the decision) \\[1ex]
\textbf{2: On-topic, but only partially fulfills the intent} \\
TThe response shows some understanding of the question’s purpose but provides limited or vague engagement. \\
\quad • Topic alignment: Yes \\
\quad • Intent alignment: Partially \\
\hangindent=1em
\hangafter=0
Example: Q: "What made this experience special?", A: "It was nice." (Sentiment expressed, but no clear reason is given) \\[1ex]
\textbf{3: Matches the topic and mostly fulfills the intent} \\
The response addresses the question appropriately and reflects a good understanding of its purpose, though some details may be missing. \\
\quad • Topic alignment: Yes \\
\quad • Intent alignment: Mostly \\
\hangindent=1em
\hangafter=0
Example: Q: "What was your goal in using this app?", A: "To reduce stress." (Clearly aligned with the question’s intent) \\[1ex]
\textbf{4: Fully aligned with both topic and intent} \\
The response directly and clearly addresses what the question is asking, fulfilling its purpose. (Detailed justification is not necessary, as long as the intent is clearly understood and responded to.) \\
\quad • Topic alignment: Yes \\
\quad • Intent alignment: Yes \\
\hangindent=1em
\hangafter=0
Example: Q: "Why did you like this product?", A: "It had a long-lasting scent and didn’t irritate my skin." (Clear, specific reason matching the question) \\[1ex]
Read the criteria carefully and follow them faithfully. \\
Return your evaluation as a JSON dictionary without any additional text: \\[1ex]
\texttt{\{} \\
\texttt{\quad "relevance": <int: 0-4>,} \\
\texttt{\quad "reason": "Detailed justification based on the criteria (2–3 sentences)"} \\
\texttt{\}} \\[1ex]

Now, let's get it started!
\texttt{Conversation: \{conversation\}} \\
\texttt{Your evaluation: \{JSON dictionary\}} \\
\specialrule{1pt}{0pt}{1pt}
\end{tabular}
\caption{Prompt used for \textit{relevance} evaluation.}
\label{tab:prompt_relevance_v4}
\end{table*}
\begin{table*}[ht] \small
\centering
\begin{tabular}{p{16cm}}
\specialrule{1pt}{0pt}{1pt}
\textbf{System Prompt:} \\
You are an expert evaluator for a human-AI conversation dataset. Your task is to evaluate the quality of the responses in the dataset based on the given criteria. \\
\specialrule{0.4pt}{1pt}{1pt}
\textbf{User Prompt:} \\
Rate how completely the response fulfills the informational requirements implied by the question. \\
Use the following \textbf{0–4 scale}. The score must be an integer. Base your judgment on whether the response addresses \textbf{all relevant parts of the question}, and \textbf{the depth or adequacy of the information provided}. \\[1ex]

Short answers like “yes” or “no” should not be automatically scored as 1: \\
\quad • If they clearly and directly address the question but lack elaboration → score 1 (Minimally Fulfilled) \\
\quad • If they are vague, off-topic, or do not engage with the question’s intent → score 0 (Not Fulfilled) \\[1ex]
\textbf{Important}: Even short or concise responses can receive a 3 (Mostly Fulfilled) or 4 (Fully Fulfilled) if they meaningfully and precisely address all parts of the question. \\[1ex]
For multi-part questions, check if each part is answered. Responses that leave parts unaddressed or only partially answered should receive lower scores. \\ \\

\textbf{0: Not Fulfilled} \\
\quad • Response does not meaningfully address the question or is a placeholder, irrelevant, or nonsensical. \\
\quad e.g., “fdsa”, “I don’t know”, “What?”, “That’s private.” \\[1ex]

\textbf{1: Minimally Fulfilled} \\
\quad • Mentions the topic but gives no substantive or relevant detail. \\
\quad • Typically applies to yes/no answers with no elaboration. \\
\quad e.g., “Movies are fun”, “I like it”, “Yes”, “No” \\[1ex]

\textbf{2: Partially Fulfilled} \\
\hangindent=1em
\hangafter=0
• Response addresses only one part of a multi-part question, or answers a single-part question incompletely. \\
\quad e.g., \\
\qquad Q: “What’s your favorite movie and why?” → A: “Inception.” \\
\qquad Q: “What did you like and dislike?” → A: “I liked it.” (missing ‘dislike’) \\[1ex]

\textbf{3: Mostly Fulfilled} \\
\quad • All parts of the question are touched on, but detail is limited or vague. \\
\quad e.g., \\
\qquad Q: “What’s your favorite movie and why?” → A: “Inception, it’s cool.” \\
\qquad Q: “What do you like and dislike?” → A: “I liked the packaging. Didn’t like the smell.” \\[1ex]

\textbf{4: Fully Fulfilled} \\
\quad • Every part of the question is clearly and fully answered, with relevant detail or reasoning. \\
\quad e.g., \\
\hangindent=2em
\hangafter=0
Q: “What’s your favorite movie and why?” → A: “Inception, because I love mind-bending plots and the visuals were stunning.” \\
\hangindent=2em
\hangafter=0
Q: “What do you like and dislike?” → A: “I liked the compact size and smooth texture. I disliked how quickly it wore off.” \\[1ex]

Read the given criteria carefully and follow them faithfully. \\
Return your evaluation as a JSON dictionary without any additional text: \\[1ex]

\texttt{\{} \\
\texttt{\quad "completeness": <int: 0-4>,} \\
\texttt{\quad "reason": "Detailed justification based on the criteria (2–3 sentences)"} \\
\texttt{\}} \\[1ex]

Now, let’s get it started! \\
\texttt{Conversation: \{conversation\}} \\
\texttt{Your evaluation: \{JSON dictionary\}} \\
\specialrule{1pt}{0pt}{1pt}
\end{tabular}
\caption{Prompt used for \textit{completeness} evaluation.}
\label{tab:prompt_completeness_v4}
\end{table*}
\begin{table*}[ht] \small
\centering
\begin{tabular}{p{16cm}}
\specialrule{1pt}{0pt}{1pt}
\textbf{System Prompt:} \\
You are an expert evaluator for a human-AI conversation dataset. Your task is to evaluate the quality of the responses in the dataset based on the given criteria. \\
\specialrule{0.4pt}{1pt}{1pt}
\textbf{User Prompt:} \\
Evaluate the overall quality of the user response using a 0–4 scale (integer only). \\
This score should represent a balanced assessment based on the combined performance across the following three dimensions: \\[0.5ex]
\quad – \textbf{Effort (0–1):} Thoughtfulness, specificity, and detail \\
\quad – \textbf{Relevance (0–1):} Alignment with the topic and intent of the question \\
\quad – \textbf{Completeness (0–1):} Coverage of the informational requirements \\[1ex]

Do not simply average the three scores. Instead, weigh the strengths and weaknesses reflected in both the scores and their justifications. \\
A response that performs strongly in one area but poorly in others may warrant a moderate overall score. \\
Likewise, a consistently adequate response across all dimensions may merit a higher score than one with extremes. \\[1ex]

You will be provided with the following input: \\
\quad – effort score: \texttt{\{effort\_score\}} \\
\quad – effort reason: \texttt{\{effort\_reason\}} \\
\quad – relevance score: \texttt{\{relevance\_score\}} \\
\quad – relevance reason: \texttt{\{relevance\_reason\}} \\
\quad – completeness score: \texttt{\{completeness\_score\}} \\
\quad – completeness reason: \texttt{\{completeness\_reason\}} \\[1ex]

Scoring scale: \\
\quad \textbf{0} – Very Poor \\
\quad \textbf{1} – Poor \\
\quad \textbf{2} – Acceptable \\
\quad \textbf{3} – Good \\
\quad \textbf{4} – Excellent \\[1ex]

Your reason should briefly explain: \\
(1) the strongest dimension, \\
(2) the weakest dimension, and \\
(3) how this balance supports your final score. \\[1ex]

Return your evaluation as a JSON dictionary with no additional text: \\[1ex]
\texttt{\{} \\
\texttt{\quad "overall\_quality": <int: 0–4>,} \\
\texttt{\quad "reason": "Justification based on the criteria (2–3 sentences)"} \\
\texttt{\}} \\[1ex]

Now, let's get it started! \\
\texttt{Conversation: \{conversation\}} \\
\texttt{Your evaluation: \{JSON dictionary\}} \\
\specialrule{1pt}{0pt}{1pt}
\end{tabular}
\caption{Prompt used for \textit{overall quality} Aggregation.}
\label{tab:prompt_overall_quality}
\end{table*}

\label{sec:appendix}

\end{document}